\documentclass{article}

\PassOptionsToPackage{numbers, compress}{natbib}

\usepackage[final,main]{neurips_2025}
\usepackage{tikz}
 \usepackage{pgfplots}
 \usepackage{array} 
\pgfplotsset{compat=1.18}
\usetikzlibrary{arrows.meta,positioning,calc,fit}

\usepackage[utf8]{inputenc} 
\usepackage[T1]{fontenc}    
\usepackage{hyperref}       
\usepackage{url}            
\usepackage{booktabs}       
\usepackage{amsfonts}       
\usepackage{nicefrac}       
\usepackage{microtype}      
\usepackage{xcolor}         
\usepackage{amsmath}
\usepackage{tikz}
\usetikzlibrary{arrows.meta,positioning,fit,calc}
\usepackage[table]{xcolor}
\usepackage{array}

\newlength{\numcolw}
\newcommand{\scorecell}[1]{%
  \ifdim #1pt > .89pt
    \cellcolor{green!30}{#1}%
  \else
    \cellcolor{red!15}{#1}%
  \fi
}\newcolumntype{R}{>{\centering\arraybackslash\columncolor{red!25}}p{\numcolw}}   

\newcolumntype{L}[1]{>{\raggedright\arraybackslash}p{#1}}
\newcolumntype{B}{>{\columncolor{blue!15}}c}    
\newcolumntype{O}{>{\columncolor{yellow!30}}c}  

\title{Delayed Attention Training Improves Length Generalization in Transformer–RNN Hybrids}

\author{%
 \quad Buu Phan\textsuperscript{\textnormal{1}}\thanks{Work done during internship at Qualcomm AI Research} 
 \quad Reza Ebrahimi
\textsuperscript{\textnormal{2}}
 \quad Sanjay Haresh
\textsuperscript{\textnormal{2}}
 \quad Roland Memisevic\textsuperscript{\textnormal{2}} \\
    \textsuperscript{1}University of Toronto \quad 
    \textsuperscript{2}Qualcomm AI Research \thanks{Qualcomm AI Research is an initiative of Qualcomm Technologies, Inc.} \\
    \texttt{truong.phan@mail.utoronto.ca}, \texttt{\{ebrahimi,sanjayh,rmemisev\}@qti.qualcomm.com}
}

\begin{document}

\maketitle

\vspace{-15pt}
\begin{abstract}
  We study length generalization in sequence models on a composite problem involving both state tracking and associative recall. Prior work finds that recurrent networks handle state tracking well but struggle with recall, whereas Transformers excel at recall yet fail to extend state-tracking capabilities to longer sequences. Motivated by the complementary strengths of these architectures, we construct hybrid models integrating recurrent and attention-based components, and train them on the combined task to evaluate whether both capabilities can be preserved. Our results reveal that, in such hybrids, the Transformer component tends to exploit shortcut solutions, leading to poor length generalization. We identify this shortcut reliance as a key obstacle and propose a simple yet effective training strategy—delaying the training of the attention layers—that mitigates this effect and significantly improves length generalization performance. Our experiments show that this approach enables hybrid models to achieve near-perfect accuracy ( $>90\%$) on hybrid sequences three times longer than those used during training.
\end{abstract}

\vspace{-12pt}
\section{Introduction}
\vspace{-5pt}

Transformers play a central role in modern language models \cite{dubey2024llama,vaswani2017attention}. Part of this success has been attributed to their ability to perform a kind of \emph{content-based retrieval}, making them especially effective for associative recall and related reasoning tasks prevalent in language modeling \cite{arora2024zoology}. 
On the other hand, transformers are known to struggle with tasks that require \emph{state tracking} and to length-generalize in the absence of carefully designed task-specific features or reasoning formats \cite{ebrahimiyour,zhou2024transformers}. 
The reason is that the lack of an inductive bias towards recurrent (step-by-step) processing allows 
transformers to rely on shortcut heuristics \cite{geirhos2020shortcut} (e.g., solving modulo addition without maintaining sequence-level state), leading to brittle length generalization behavior. 
In contrast, recurrent models such as LSTMs \cite{hochreiter1997long} naturally maintain state across long sequences and generalize well on state-tracking tasks, but their fixed-capacity memory makes content-based retrieval challenging for these models. 

\textbf{Contributions. } Given the importance of length generalization for general reasoning, we investigate whether a hybrid architecture can effectively combine the complementary strengths of recurrent and attention-based models. To this end, we design a synthetic task that jointly evaluates a model’s ability to perform both \emph{state tracking} and \emph{selective recall}. Our experiments reveal that transformers are prone to adopting shortcut solutions, even within hybrid architectures, thereby undermining their ability to generalize over longer sequences. Nevertheless, we find that by delaying the training of the attention component, we can effectively prevent such shortcuts and enable the network to preserve state-tracking capabilities while still benefiting from the recall advantages of attention.


\vspace{-5pt}
\section{Methods and Results}
\vspace{-5pt}

\textbf{Experimental Setup. } We design each training instance as a
sequence consisting of key–value pairs, followed by two query markers that
require multiple different forms of reasoning. Each input sequence takes the form:
\(
\langle \mathrm{bos} \rangle \; k_1 v_1 \, k_2 v_2 \, \cdots \, k_n v_n \,
\langle \mathrm{modulo} \rangle \, m \,
\langle \mathrm{recall} \rangle \, k_j v_j ,
\)
where $k_i$ and $v_i$ are discrete tokens representing keys and values,
respectively. The sequence begins with a special start symbol $\langle
\mathrm{bos}\rangle$. The segment $\langle \mathrm{modulo} \rangle$ queries the
model for the sum of all values modulo 10, and the segment $\langle
\mathrm{recall} \rangle$ queries the model for the value associated with a
randomly chosen key $k_j$ from the sequence. A prediction is considered correct only if the model successfully answers both tasks.


\begin{figure}[t]
\centering
\vspace{-45pt}
\makebox[\linewidth][l]{\hspace*{-6.3em}
\begin{tabular}{@{}p{0.495\linewidth}@{\hspace{-5.em}}p{0.495\linewidth}@{}}

\begin{minipage}[t]{\linewidth}
\vspace{0pt}\centering
\begin{tikzpicture}[
  font=\tiny, >=Latex, node distance=12mm,
  box/.style={draw, rounded corners=2pt, minimum width=12mm, minimum height=2mm, align=center, fill=blue!5},
  sub/.style={draw, rounded corners=2pt, minimum width=12mm, minimum height=2mm, align=center, fill=green!10},
  add/.style={draw, circle, inner sep=0pt, minimum size=3.5mm, fill=orange!10},
  group/.style={draw, thick, rounded corners=4pt, fill=red!30, fill opacity=0.2}
]
\node[box, fill=blue!15] (x) {Input $X$};
\node[add, above=7mm of x] (add1) {$+$};
\node[add, above=11mm of add1] (add2) {$+$};
\node[add, above=11mm of add2] (add3) {$+$};
\node[sub, fill=orange!55, above=3mm of add3] (ln) {LayerNorm};
\node[box, fill=blue!15, above=3mm of ln] (y) {Output $Y$};
\draw[->] (x) -- (add1);
\draw[->] (add1) -- (add2);
\draw[->] (add2) -- (add3);
\draw[->] (add3) -- (ln);
\draw[->] (ln) -- (y);
\coordinate (branchpoint1) at ($(x)!0.27!(add1)$);
\node[sub, right=8mm of add1, yshift=-3.4mm] (lstm1) {LSTM};
\draw[->] (branchpoint1) -| (lstm1.south);
\draw[->] (lstm1.north) |- (add1.east);
\coordinate (branchpoint2) at ($(add1)!0.16!(add2)$);
\node[sub] (mha) at ([yshift=8.1mm]lstm1.north) {MHA};
\draw[->] (branchpoint2) -| (mha.south);
\node[sub] (ln1) at ([yshift=3.2mm]mha.north) {LayerNorm};
\draw[-] (mha.north) -| (ln1.south);
\draw[->] (ln1.north) |- (add2.east);
\coordinate (branchpoint3) at ($(add2)!0.16!(add3)$);
\node[sub] (mlp) at ([yshift=12.7mm]mha.north) {MLP};
\node[sub] (ln2) at ([yshift=3.25mm]mlp.north) {LayerNorm};
\draw[->] (branchpoint3) -| (mlp.south);
\draw[-] (mlp.north) -| (ln2.south);
\draw[->] (ln2.north) |- (add3.east);
\node[group, fit=(mha)(ln2), draw=red] (delayedbox) {};
\node[above=0.mm of delayedbox, text=red!70!black, font=\scriptsize, align=center] {Removed \\ during DAT};
\end{tikzpicture}
\end{minipage}
&
\begin{minipage}[t]{\linewidth}
\vspace{15pt}\raggedright
\hspace*{-15em}
\footnotesize
\setlength{\tabcolsep}{1.6pt}
\renewcommand{\arraystretch}{1.05}

\begin{tabular}{|
 L{0.4\linewidth}|
  *{5}{B|}   
  *{7}{O|}   
}
\hline
\textbf{Model} & \textbf{5} & \textbf{10} & \textbf{15} & \textbf{20} & \textbf{25} &
\textbf{30} & \textbf{35} & \textbf{40} & \textbf{45} & \textbf{50} & \textbf{75} & \textbf{100} \\
\hline
\rowcolor{white}
\emph{Transformer} & \scorecell{1.00} & \scorecell{1.00} & \scorecell{0.99} &
 \scorecell{0.99} & \scorecell{0.99} & \scorecell{0.49} &
 \scorecell{0.10} & \scorecell{0.09} & \scorecell{0.09} &
 \scorecell{0.09} & \scorecell{0.05} & \scorecell{0.03} \\
\hline
\emph{LSTM} & \scorecell{1.00} & \scorecell{0.73} & \scorecell{0.52} & \scorecell{0.41} & \scorecell{0.35} & \scorecell{0.32} & \scorecell{0.28} & \scorecell{0.25} & \scorecell{0.23} & \scorecell{0.22} & \scorecell{0.17} & \scorecell{0.15} \\
\hline
\emph{LSTM $\circ$ Attn} & \scorecell{1.00} & \scorecell{1.00} & \scorecell{0.99} & \scorecell{0.99} & \scorecell{0.99} & \scorecell{0.45} & \scorecell{0.12} & \scorecell{0.10} & \scorecell{0.09} & \scorecell{0.09} & \scorecell{0.05} & \scorecell{0.04} \\
\emph{(+ DAT)}        &  \scorecell{1.00}& \scorecell{1.00}&  \scorecell{1.00}&  \scorecell{1.00}&  \scorecell{1.00}&  \scorecell{0.99}&  \scorecell{0.99}&  \scorecell{0.99}&  \scorecell{0.99}& \scorecell{0.99} & \scorecell{0.93} &  \scorecell{0.88}\\
\hline
\emph{Attn $\circ$ LSTM} & \scorecell{1.00} & \scorecell{1.00} & \scorecell{0.99} & \scorecell{0.99} & \scorecell{0.99} & \scorecell{0.83} & \scorecell{0.36} & \scorecell{0.18} & \scorecell{0.12} & \scorecell{0.09} & \scorecell{0.07} & \scorecell{0.05} \\
\emph{(+ DAT)}        &  \scorecell{1.00}& \scorecell{1.00}&  \scorecell{1.00}&  \scorecell{1.00}&  \scorecell{1.00}&  \scorecell{0.99}&  \scorecell{0.99}&  \scorecell{0.99}&  \scorecell{0.99}& \scorecell{0.99} & \scorecell{0.80} &  \scorecell{0.56}\\
\hline
\emph{Hybrid Block}   & \scorecell{1.00} & \scorecell{1.00} & \scorecell{1.00} & \scorecell{1.00} & \scorecell{1.00} & \scorecell{0.56} & \scorecell{0.21} & \scorecell{0.12}  & \scorecell{0.10} & \scorecell{0.10} &\scorecell{0.09}  & \scorecell{0.08} \\
\emph{(+ DAT)}        & \scorecell{1.00} & \scorecell{1.00} & \scorecell{1.00} & \scorecell{1.00} & \scorecell{1.00} & \scorecell{1.00} & \scorecell{1.00} & \scorecell{0.99} & \scorecell{0.99} & \scorecell{0.99} & \scorecell{0.93} & \scorecell{0.85} \\
\hline
\emph{LSTM $\circ$ Attn $\circ$ LSTM}  &  \scorecell{1.00}&  \scorecell{1.00}&  \scorecell{1.00}& \scorecell{1.00} &  \scorecell{1.00} &\scorecell{ 0.52} &  \scorecell{0.13}&  \scorecell{0.05}& \scorecell{0.05}&  \scorecell{0.07}&  \scorecell{0.06}&  \scorecell{0.06}\\
\emph{(+ DAT)}        &  \scorecell{1.00}& \scorecell{1.00} & \scorecell{1.00} & \scorecell{1.00} &  \scorecell{1.00}&  \scorecell{1.00}& \scorecell{0.99} & \scorecell{0.99} &  \scorecell{0.99}&  \scorecell{0.99}&  \scorecell{0.91}&  \scorecell{0.80}\\
\hline
\emph{LSTM + Attn}   & \scorecell{1.00} & \scorecell{1.00} & \scorecell{1.00} & \scorecell{1.00} & \scorecell{1.00} & \scorecell{0.28} & \scorecell{0.20} & \scorecell{0.19} & \scorecell{0.18} & \scorecell{0.16} & \scorecell{0.12} & \scorecell{0.08} \\
\emph{(+ DAT)}        & \scorecell{1.00} & \scorecell{1.00} & \scorecell{1.00} & \scorecell{1.00} & \scorecell{1.00} & \scorecell{1.00} & \scorecell{1.00} & \scorecell{0.99} & \scorecell{0.99} & \scorecell{0.99} & \scorecell{0.95} & \scorecell{0.85} \\
\hline
\end{tabular}

\end{minipage}
\\
\end{tabular}
}
\vspace{-2pt}
\caption{Left: Hybrid block diagram with a single attention layer, i.e. the red box, which is replicated $\times 4$ in experiments. 
Right: Accuracy across sequence lengths, evaluated on both in-distribution $(\leq 25)$ and out-of-distribution $(>25)$ cases. Accuracy with $>90\%$ is highlighted in green color.}

\vspace{-15pt}
\label{main_fig}
\end{figure}
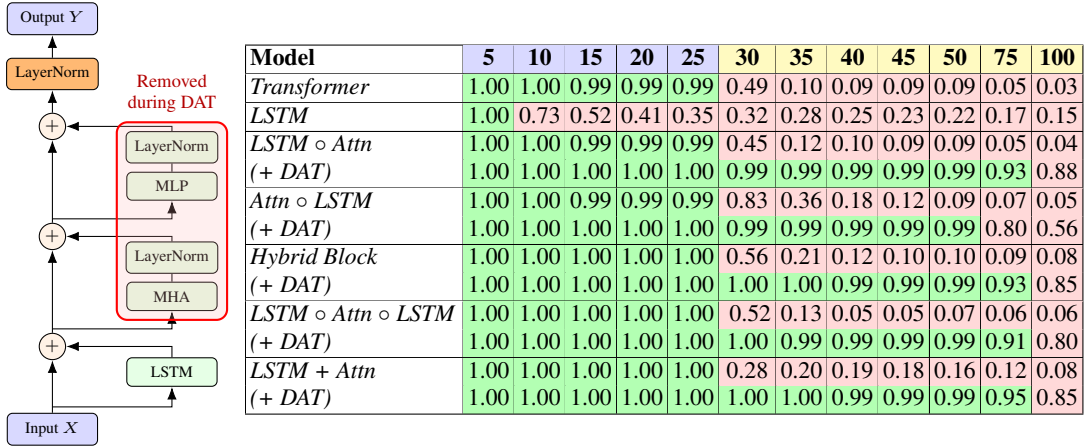

\textbf{Architectures. }We fix the embedding dimension to \(d=256\) for all models and adopt 
RoPE~\cite{su2024roformer} as the default positional encoding method. 
We evaluate two baselines—Transformer-only and LSTM-only—along with four hybrid 
variants composed of the same components. Details of each hybrid are provided in 
Appendix~\ref{arch}. Figure~\ref{main_fig} (left) illustrates one such design, 
referred to as the \emph{Hybrid Block}, where an LSTM block is integrated into 
the residual architecture of a Post-LayerNorm Transformer.

\textbf{Delayed Attention Training (DAT).} 
 In the initial phase of DAT, we deactivate the attention modules, effectively disabling their contribution while the
recurrent pathway learns to handle state tracking. After a fixed number of
epochs, we enable gradient flow through the attention layers, allowing them
to train jointly with the recurrent module. This encourages the model to
first establish a robust state-tracking mechanism before leveraging
attention for content-based retrieval. Figure 1 (left) shows how we deactivate the attention function in the hybrid block during early phase.

\textbf{Results. } Figure~\ref{main_fig} (right) compares the performance of mentioned architectures in the proposed task. LSTM underperforms even in-distribution due to its restricted memory capacity in handling the recall component. 
Transformers performs relatively well in-distribution but fail to perform state-tracking and consequently do not length generalize. 
Hybrid architectures also achieve high accuracy in-distribution, but their ability to generalize degrades with increasing sequence length, as the incorporation of attention tends to encourage shortcut learning. 
This effect is most clearly observed when training solely on the modulo-10 task (without the recall component) (see Appendix \ref{additional}): while an LSTM model exhibits perfect length generalization, augmenting it with an additional attention layer immediately forfeits this capability and results in rapid performance degradation. 
This is somewhat counterintuitive, since one might naturally expect the attention mechanism to merely propagate the hidden states to the final projection head without impairing generalization.  Importantly, DAT mitigates this failure and demonstrate robust length generalization across different hybrid architectures.  Finally, we note that the accuracy drop at sequence length 100 is due to the reduced sharpness of the attention softmax\cite{velickovic2410softmax}, evidenced in Table \ref{recall_only_result} in Appendix \ref{additional}.

\vspace{-8pt}
\section{Conclusion}
\vspace{-5pt}

In this work, we studied the challenge of length generalization on tasks that require both recall and state tracking. We showed that while LSTMs generalize well on arithmetic state-tracking tasks and Transformers excel at selective recall, naïve hybrid architectures, which combine the two, fail, since the attention mechanism allows a model to learn shortcuts. 
To address this, we proposed \emph{Delayed Attention Training}, a simple strategy that prevents shortcut learning and enables reliable length generalization in hybrid models. Our experiments demonstrate that this approach recovers the complementary strengths of recurrence and attention. Looking forward, extending the framework to large-scale language modeling and designing specialized modules or training curricula may further improve generalization in real-world settings.

\newpage
\bibliographystyle{plainnat} 
\bibliography{reference} 

\newpage 
\appendix

\section{Hybrid Architectures} \label{arch}
\textbf{Training Setup. }We use Adam optimizer with the learning rate of $10^{-4}$ for the first 500 epoches and $10^{-5}$ afterwards for all end-to-end setup. For DAT setup, we train the LSTM component for 1500 epoches with learning rate of $10^{-4}$. During attention training phase, we use the learning rate of $10^{-4}$ for 500 epoches before switching to $10^{-5}$. Each epoch consists of $2^{14}$ randomly generated samples. 

\textbf{Architectures. } We describe the hybrid architectures in details as follows. Unless otherwise stated, we use 4 block in the transformer components with post-layernorm setup.

\emph{(1) Attention-on-LSTM ({Attn $\circ$ LSTM}).} An LSTM encodes the sequence, and its hidden states are fed to a self-attention (MHA+MLP) block: \(H=\mathrm{LSTM}(X),\; Z=\mathrm{Attn}(H)\).

\emph{(2) LSTM-on-Attention ({LSTM $\circ$ Attn}).} A self-attention block first produces contextualized features that are then processed by an LSTM: \(H=\mathrm{Attn}(X),\; Z=\mathrm{LSTM}(H)\).

\emph{(3) Parallel branch-and-add  (LSTM}\(+\)\emph{Attn).} LSTM and self-attention process the input in parallel and their outputs are combined by an elementwise sum: \(H_{\text{lstm}}=\mathrm{LSTM}(X),\; H_{\text{attn}}=\mathrm{Attn}(X),\; Z=H_{\text{lstm}}+H_{\text{attn}}\).

\emph{(4) Hybrid block (Fig.\ref{main_fig}).} A composite module that integrates LSTM, self-attention, and MLP within a single residual block as depicted in Figure~1 (Left).

\emph{(5) Sandwich Attention (\emph{LSTM $\circ$ Attn $\circ$ LSTM}).} The sequence is encoded by an LSTM, passed through a self-attention block (MHA+MLP), and then processed by another LSTM: $H=\mathrm{LSTM}(X),\; Z=\mathrm{Attn}(H),\; Y=\mathrm{LSTM}(Z)$.

\paragraph{Remarks.} Beyond LSTMs, there exist several alternative sequence models based on state-space formulations, such as Mamba \cite{gu2023mamba} and its hybrid variants, e.g., Hymba \cite{dong2025hymba}. Since state-space models are known to lack effective state-tracking capabilities \cite{ebrahimi2025revisiting}, we exclude them from our hybrid design, expecting the conclusions drawn for transformers in Appendix \ref{additional} to extend naturally to these models.

\section{Additional Results} \label{additional}
\begin{table}
    \centering
    \footnotesize
    \begin{tabular}{|
 L{0.20\linewidth}|
  *{5}{B|}   
  *{7}{O|}   
}
\hline
\textbf{Model} & \textbf{5} & \textbf{10} & \textbf{15} & \textbf{20} & \textbf{25} &
\textbf{30} & \textbf{35} & \textbf{40} & \textbf{45} & \textbf{50} & \textbf{75} & \textbf{100} \\
\hline
\rowcolor{white}
\emph{Transformer} & \scorecell{1.00} & \scorecell{1.00} & \scorecell{1.00} &
 \scorecell{1.00} & \scorecell{1.00} & \scorecell{1.00} &
 \scorecell{1.00} & \scorecell{1.00} & \scorecell{0.99} &
 \scorecell{0.99} & \scorecell{0.90} & \scorecell{0.80} \\
\hline
\emph{LSTM} & \scorecell{1.00} & \scorecell{0.98} & \scorecell{0.95} & \scorecell{0.89} & \scorecell{0.83} & \scorecell{0.76} & \scorecell{0.70} & \scorecell{0.64} & \scorecell{0.59} & \scorecell{0.53} & \scorecell{0.37} & \scorecell{0.28} \\
\hline
\emph{LSTM $\circ$ Attn} &  \scorecell{1.00} &  \scorecell{1.00} &  \scorecell{1.00} &  \scorecell{1.00} &  \scorecell{1.00} &  \scorecell{0.99} & \scorecell{0.99}&  \scorecell{0.99}&  \scorecell{0.99}&  \scorecell{0.99}& \scorecell{0.90}&  \scorecell{0.76}\\
\hline
\emph{Attn $\circ$ LSTM} & \scorecell{1.00} & \scorecell{1.00} & \scorecell{1.00} & \scorecell{1.00} & \scorecell{1.00} & \scorecell{1.00} & \scorecell{0.99}& \scorecell{0.99} & \scorecell{0.99} & \scorecell{0.99} & \scorecell{0.96} & \scorecell{0.89} \\
\hline
\emph{Hybrid Block}  & \scorecell{1.00} & \scorecell{1.00} & \scorecell{1.00}& \scorecell{1.00} &  \scorecell{1.00}  &  \scorecell{1.00} &  \scorecell{1.00} &   \scorecell{1.00} &   \scorecell{0.99}& \scorecell{0.99} &\scorecell{0.98} &\scorecell{0.92}  \\
\hline
\emph{LSTM $\circ$ Attn $\circ$ LSTM}  & \scorecell{1.00} & \scorecell{1.00} & \scorecell{1.00} & \scorecell{1.00}& \scorecell{1.00} & \scorecell{1.00}& \scorecell{1.00}& \scorecell{0.99} & \scorecell{0.99} & \scorecell{0.99} & \scorecell{0.97}& \scorecell{0.90} \\
\hline
\emph{LSTM + Attn}   & \scorecell{1.00} & \scorecell{1.00} & \scorecell{1.00} & \scorecell{1.00} & \scorecell{1.00} & \scorecell{1.00} & \scorecell{1.00} & \scorecell{1.00} & \scorecell{0.99} & \scorecell{0.99} & \scorecell{0.92} & \scorecell{0.82} \\
\hline
\end{tabular}
\vspace{10pt}
    \caption{Results on the Recall-Only setting.}
    \label{recall_only_result}
\end{table}

\begin{table}
    \centering
    \footnotesize
    \begin{tabular}{|
 L{0.20\linewidth}|
  *{5}{B|}   
  *{7}{O|}   
}
\hline
\textbf{Model} & \textbf{5} & \textbf{10} & \textbf{15} & \textbf{20} & \textbf{25} &
\textbf{30} & \textbf{35} & \textbf{40} & \textbf{45} & \textbf{50} & \textbf{75} & \textbf{100} \\
\hline
\rowcolor{white}
\emph{Transformer} &\scorecell{1.00}  & \scorecell{1.00} & \scorecell{1.00} & \scorecell{0.99}
  &\scorecell{0.98}  & \scorecell{0.51} & \scorecell{0.11}
  & \scorecell{0.10} & \scorecell{0.10} & \scorecell{0.10}
  & \scorecell{0.10} & \scorecell{0.10} \\
\hline
\emph{LSTM} & \scorecell{1.00} & \scorecell{1.00} & \scorecell{1.00} & \scorecell{1.00} & \scorecell{1.00} & \scorecell{1.00} & \scorecell{1.00} & \scorecell{1.00} & \scorecell{1.00} & \scorecell{1.00} & \scorecell{1.00} & \scorecell{1.00} \\
\hline
\emph{LSTM $\circ$ Attn} & \scorecell{1.00}& \scorecell{1.00} &\scorecell{1.00}  &\scorecell{1.00}  &\scorecell{1.00}  & \scorecell{0.20} & \scorecell{0.10} & \scorecell{0.10} & \scorecell{0.10}& \scorecell{0.10}& \scorecell{0.10}& \scorecell{0.10}\\
\hline
\emph{Attn $\circ$ LSTM} & \scorecell{1.00}&  \scorecell{1.00}&  \scorecell{1.00}&  \scorecell{1.00}&  \scorecell{1.00}& \scorecell{0.63} &  \scorecell{0.10}& \scorecell{0.10} & \scorecell{0.10} & \scorecell{0.10} & \scorecell{0.10} & \scorecell{0.10} \\
\hline
\emph{Hybrid Block}   & \scorecell{1.00}& \scorecell{1.00} & \scorecell{1.00} & \scorecell{1.00} & \scorecell{1.00} &  \scorecell{0.45} &   \scorecell{0.22}&  \scorecell{0.13} &  \scorecell{0.10} &  \scorecell{0.10} &  \scorecell{0.10} &  \scorecell{0.10} \\
\hline
\emph{LSTM $\circ$ Attn $\circ$ LSTM}  &  \scorecell{1.00}& \scorecell{1.00}& \scorecell{1.00}& \scorecell{1.00}&  \scorecell{1.00} & \scorecell{0.73}& \scorecell{0.16} & \scorecell{0.08} & \scorecell{0.09}&  \scorecell{0.10}& \scorecell{0.10} &  \scorecell{0.10}\\
\hline
\emph{LSTM + Attn}   &  \scorecell{1.00}&  \scorecell{1.00}& \scorecell{1.00}&  \scorecell{1.00}&  \scorecell{0.99}& \scorecell{0.11} & \scorecell{0.11} & \scorecell{0.10} & \scorecell{0.10} & \scorecell{0.10} &  \scorecell{0.10}& \scorecell{0.10} \\
\hline
\end{tabular}
\vspace{10pt}
    \caption{Results on the Modulo-Only setting.}
    \label{modulo_only_result}
    \vspace{-20pt}
\end{table}

We report results on the performance of each model under single-task settings, where models are trained and evaluated solely on either the recall or modulo task in Table \ref{recall_only_result} and \ref{modulo_only_result}. The results confirm that the transformer component enables all hybrid models to excel at the recall task but leads to poor length generalization on the state-tracking task.


\newpage

\end{document}